\documentclass[letterpaper, 10 pt, conference]{ieeeconf}  

\IEEEoverridecommandlockouts                              

\usepackage{graphics} 
\usepackage{epsfig} 
\usepackage{amsmath} 
\usepackage{amssymb}  
\usepackage{color}
\usepackage{xcolor}
\usepackage{gensymb}

\usepackage{diagbox}

\title{\LARGE \bf
Transfer from Simulation to Real World through\\Learning Deep Inverse Dynamics Model
}

\author{Paul Christiano, Zain Shah, Igor Mordatch, Jonas Schneider, \\Trevor Blackwell, Joshua Tobin, Pieter Abbeel, and Wojciech Zaremba
\\OpenAI, San Francisco, CA, USA}

\begin{document}

\maketitle
\thispagestyle{empty}
\pagestyle{empty}

\begin{abstract}
Developing control policies in simulation is often more practical and safer than directly running experiments in the real world.  This applies to policies obtained from planning and optimization, and even more so to policies obtained from reinforcement learning, which is often very data demanding.  However, a policy that succeeds in simulation often doesn’t work when deployed on a real robot.  Nevertheless, often the overall gist of what the policy does in simulation remains valid in the real world.  In this paper we investigate such settings, where the sequence of states traversed in simulation remains reasonable for the real world, even if the details of the controls are not, as could be the case when the key differences lie in detailed friction, contact, mass and geometry properties.  During execution, at each time step our approach computes what the simulation-based control policy would do, but then, rather than executing these controls on the real robot, our approach computes what the simulation expects the resulting next state(s) will be, and then relies on a learned deep inverse dynamics model to decide which real-world action is most suitable to achieve those next states.  Deep models are only as good as their training data, and we also propose an approach for data collection to (incrementally) learn the deep inverse dynamics model.  Our experiments shows our approach compares favorably with various baselines that have been developed for dealing with simulation to real world model discrepancy, including output error control and Gaussian dynamics adaptation.  
\end{abstract}

\section{Introduction}
Many methods exist for generating control policies in simulated environments, including methods based on motion planning, optimization, control, and learning.   However, an important practical challenge is that often there are discrepancies between simulation and the real world, which results in policies that work well in simulation yet perform poorly in the real world. 

Significant bodies of work exist that strive to address this challenge.  One important line of work studies how to improve simulators to better match reality, which involves improving simulation of contact, non-rigidity, friction, as well as improving identification of physical quantities needed for accurate simulation such as mass, geometry, friction coefficients, elasticity.  However, despite significant progress, discrepancies continue to exist, and more accurate simulation can have the downside of being slower. 

Another important line of work studies robustness of control policies, which could be measured through, for example, gain and phase margins, and robust control methods exist that can optimize for these.  Optimizing for robustness means finding control policies that apply across a wide range of possible real worlds, but unfortunately tends to come at the expense of performance in the one specific real world the system is faced with.

Adaptive methods, which is the topic of this paper, do not use the same policy for the entire family of possible environments, but rather try to learn about the specific real world the system is faced with.  In principle, such methods can exploit the physics of the real world and behave in the optimal way.

Concretely, our work considers the following problem setting: We assume to be given a simulator and a method for generating policies that perform well in simulation.  The goal is to leverage this to perform well in new real-world situations.  To achieve this, a training period exists during which an adaptation mechanism can be trained to learn to adapt from simulation to real world by collecting experience on the real system, but without having access to the new real-world situations that the system will be evaluated on later.


We leverage the following intuition: Often policies found from simulation capture the high-level gist well (e.g., overall trajectory), but fail to accurately capture some of the lower-level details, such as friction, stiction, backlash, hysteresis, precise measurements, precise deformation, etc.  Indeed, this is the type of situation that motivates the work in this paper and in which we will be evaluating our approach (as well as baselines).

Note that while we assume that a method exists for generating policies in simulation, our approach is agnostic to the details of this method, which could be based on any techniques from motion planning, optimization, control, learning, and others, which return a policy, which could be a model-predictive policy which uses the simulator in its inner loop.

Our approach proceeds as follows:  During execution on a test trajectory, at each time step it computes what the simulation-based control policy would do, but then, rather than executing these controls on the real robot, our approach computes what the simulation expects the next state(s) will be, and then relies on a learned deep inverse dynamics model to decide which real-world action is most suitable to achieve those next states.   As our experiments show, when these inverse dynamics models are trained on sufficient data, this results in compelling transfer from simulation to real world, in particular with challenging dynamics involving contact and collision.  To collect the training data, there is a training phase which proceeds the same way, but only has access to a poor inverse dynamics model, and then uses the collected data to improve the model.  Our experiments show that having the training data collection be similar to the test time conditions improves results significantly compared to data collection based on just applying random controls.   To maximize data collection efficiency, target trajectories for training are initially short (or cut short once significantly deviating from the target).

Our experiments validate the applicability of our approach through two families of experiments: (i) Sim1 to Sim2 Transfer: To better understand the transfer capabilities, we first study transfer from one simulation (Sim1) to another simulation (Sim2).  We consider several standard tasks: Reacher, Hopper, Cheetah, Humanoid from MuJoCo / OpenAI Gym ~\cite{mujoco} ~\cite{brockman2016openai}.  For each experiment Sim2 has the same type of robot as Sim1, but the physical properties are different (change in mass, link lengths, friction coefficients, torque scale and limits).  (ii) Sim to Real Transfer with Fetch: In this family of experiments we study transfer of policies that work well for a simulated Fetch robot onto a real Fetch robot.  To calibrate performance, we consider as a baseline a PD controller tuned for our Fetch robot.

We compare our approach with output error control \cite{ljung99} and Gaussian Dynamics Adaptation \cite{fu15}, two established approaches to handle mismatch between simulation and real world.

\section{Related work}

\begin{figure*}[!ht]
  \centering
  \includegraphics[width=1.0\linewidth]{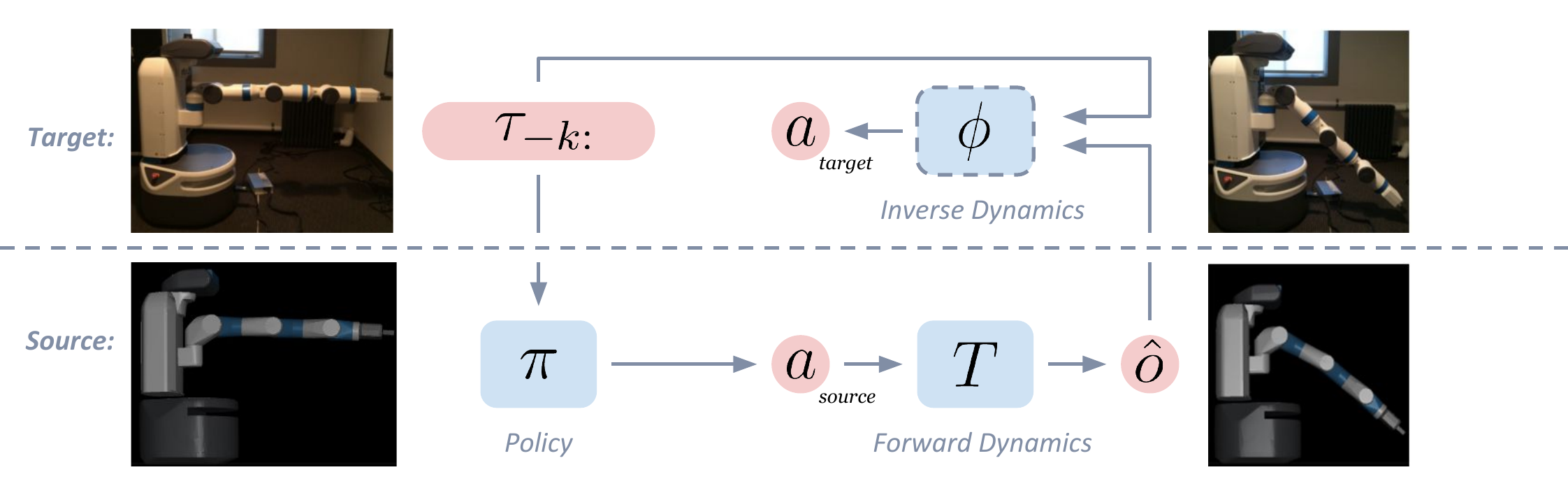}
\caption{Overview of our method applied to Fetch robot in source simulator (bottom) and target physical world (top). Given an existing source domain policy $\pi$ and forward dynamics $T$, we learn inverse dynamics neural network $\phi$ and use it to generate actions $a_\text{target}$ for the physical Fetch robot at any given time instant.}
\label{fig:overview}
\end{figure*}

Simulation has been an invaluable tool in advancing the development of robotics and many simulation techniques have been developed over the years. Reduced coordinate rigid multibody dynamics are especially suited for simulating articulated robots \cite{featherstone07}. Unfortunately, many significant physical effects may not be possible to model with such simulation approaches. Flexible or inflatable bodies \cite{tan12} \cite{eppner16}, area contact \cite{gilardi13}, interaction with fluids \cite{tan11} \cite{pan16} are just a few of such examples. More accurate simulators, such as those based on Finite Element Method \cite{abaqus02} can be used to more closely match such real world effects, but they can be extremely computationally intensive (requiring days to compute seconds of simulation) and furthermore can be numerically ill-conditioned, which makes them difficult to use within numerical trajectory or policy optimization methods. Our method allows the use of simple, high-performance, and numerically smooth rigid body simulators (we use MuJoCo \cite{mujoco}) for policy or trajectory optimization, while still being able to adapt to complex effects present in the real world.

Even if a simulator were capable of modeling all the physical effects of interest, it would still require detailed and accurate model parameters (such as mass distributions, material properties, etc.). A significant body of research has focused on identifying these parameters from observations of robots' behavior in the real world, but tend to require separate specialized identification approaches and models for different robot platforms, such as legged robots \cite{levashov12}, helicopters \cite{kanade99}, or fixed-wing UAVs \cite{uav12}. Furthermore, individual physical effects also require specialized expert-designed models and parameter identification methods, such as motor backlash \cite{hovland02}, hydraulic actuation \cite{boulet92}, series elastic actuation \cite{sentis15}, or pneumatic actuation \cite{tassa13}. Our learned deep inverse dynamics models are based on past histories of observed states and in principle have the ability to model the above effects and platforms in one simple unified method without requiring any domain-specific manual model design and identification.  

To remove the need for explicit dynamics, learning of dynamics models has received much attention in recent years. A number of approaches learn forward dynamics models - functions mapping current state and action to a next state \cite{ljung99} \cite{punjani15}. Such functions can then be used to solve for actions that lead to desired next state. Alternatively, inverse dynamics models learn a mapping from current and next state to an action that achieves the transition between the two \cite{peters10}, \cite{peters15}, \cite{franzi16}. Such models are appealing because their output can be directly used for control, and is the model type we use in this work. The data for model learning is typically gathered in a batch fashion, either from random trajectories, or from representative demonstrations. This can be problematic if the robot state trajectories resulting from policy execution do not match the model training trajectories. An alternative is to learn dynamics models in an on-line fashion, constantly adapting the model based on an incoming stream of observed states and actions \cite{fu15} \cite{mordatch16} \cite{yip2014} \cite{deepmpc15}. These approaches however are slow to adapt to rapidly-changing dynamics modes (such as those arising when making or breaking contact) and may be problematic when applied on robots performing rapid motions. Another alternative is to iteratively intertwine data collection and dynamics model learning \cite{pilco11} \cite{finn16}. Such approaches concentrate training data in the regions of the state space that are relevant for task completion and inspire the data collection procedure in our work. 

A number of options are available for representation of learned dynamics functions, from linear functions \cite{mordatch16} \cite{yip2014}, to Gaussian processes \cite{reidmiller14} \cite{ko09} \cite{pilco11}, to deep neural networks \cite{punjani15} \cite{fu15}. Linear functions are very efficient to evaluate and solve controls for, but have limited expressive power. Gaussian Processes are able to provide model uncertainty estimates, but are problematic to scale to large dataset sizes. Deep neural networks are an expressive class of functions independent of dataset size and are what we use in this work. 

Our approach to transfer between simulator and the real world is based on adapting actions. There is a rich body of work focusing on adapting policies, rather than actions in the context of reinforcement learning \cite{taylor2009transfer} \cite{barrett10} \cite{cutler2015real}. Another alternative is to consider robust control methods in simulation that produce policies that are robust to mismatch between simulator and the real world \cite{zhou98} \cite{mordatch15}. In addition to actions, adaptation of states and observations between simulation and the real world is another challenging problem \cite{tzeng15} \cite{hoffman2013efficient} \cite{devin15} \cite{donahue2014decaf}. In the current work, we choose to focus solely on adaptation of actions and leave other types of adaptation for future work.

\section{Method}
\subsection{Setting}
\label{secc:method_setting}

We study transfer from a source environment to a
target environment.  Typically the source environment would be a
simulator, and the target environment would be a physical robot.
However, in order to validate our method we start by having simulator
both in the source and in the target domain. This setup has merit in developing an experimental understanding of our approach, as we can control the degree of variation between source and target environments.
Our final experiments are in transfer from a simulator to the physical robot.

For each environment we denote the state space by $S$, the action
space by $A$ and the observation space by $O$.  
Points $s \in S$, $a \in A$, $o \in O$
are states, actions, and observations.  The state is not assumed
observed.  Overloading notation slightly, the agent makes noisy and incomplete observations of the underlying system,
$o = o(s) \in O$,
which typically don't expose some latent factors (e.g., fluctuating temperature or motor backlash).
The special situation where the state is observed is readily
captured by having the observation function $o(s) = s$.
The system forward dynamics are given by a function from state-action pair to a new state:
$T(s, a) = s'$. 

We use subscripts to explicitly distinguish between the source
environment and the target environment.  For example, $A_{\rm
  source}$ denotes the action space in the source environment, and
$A_{\rm target}$ denotes the action space in the target environment.

A trajectory $\tau$ is a sequence of
observations and actions: $(o_1, a_1, o_2, a_2, \dots)$.
We write $\tau_{H:H+k}$ to refer to the subsequence $(o_H, a_H, \ldots, o_{H+k-1}, a_{H+k-1}, o_{H+k})$.
We write $\tau_{-k:}$ to refer to the most recent $k$ observations and $k-1$ actions in a trajectory,
and $\tau_{-1}$ to refer to the most recent observation.

A policy $\pi$ is a mapping from 
observations to actions, that depends on the last $k$ observations, prescribing $a = \pi(\tau_{-k:})$. Our goal is to find a policy $\pi_{\rm target}$ that
performs well in the target environment.  

Rather than learning a policy for the target environment from scratch,
we assume that we have access to a competent policy $\pi_{\rm source}$
in the source environment.  Such policy could be obtained
  through any of a variety of methods, including motion planning,
  model-predictive or optimization-based control, reinforcement
  learning, etc.  Our approach is agnostic to how the policy $\pi_{\rm
    source}$ was
  obtained.

\subsection{Transfer to the target environment}
\label{secc:method_transfer}

Rather than directly
executing $\pi_{\rm source}$ in the target environment, we seek to
transfer the high-level properties of $\pi_{\rm source}$ to be re-used in the
target environment, but not its lower-level specifics. Our approach is illustrated in  Figure~\ref{fig:overview}.  During execution, we repeat the following at every time instant: consider the recent history
of observations $\tau_{-k:}$, compute the action $a_{\rm source} = \pi_{\rm
  source}(\tau_{-k:})$ which our source policy  prescribes for the
source environment.  Simulate what observation $\hat{o}_{\rm next} = o(T_{\rm
  source}(\tau_{-k:}, a_{\rm source}))$ would be attained at the next time step in the source environment, and then
compute $a_{\rm target} = \phi(\tau_{-k:}, \hat{o}_{\rm next})$. $\phi$ is a learned inverse dynamics model for the target
environment, which takes in the recent history of actions and
observations, as well as the desired next observations, and produces the action in the target domain that leads as close as possible to the desired observation $\hat{o}_{\rm next}$.

Putting this all together, we have:
$$\pi_{\rm target}(\tau_{-k:}) = 
\phi(\tau_{-k:}, o(T_{\rm source}(\tau_{-k:}, \pi_{\rm source}(\tau_{-k:})))).$$

To be able to execute this approach, we assume that the simulator
provides a forward dynamics model $T_{\rm source}$ that allows us to compute a reasonable estimate of the next state $s'$ and observation $o(s')$.

If the learned inverse dynamics model is sufficiently accurate,
then the next observation $o_{\rm target}$ after taking action $\pi_{\rm target}(\tau_{-k:})$
will be similar to $\hat{o}_{\rm next}$.

For this approach to be meaningful, it is assumed that
source and target environments have the same actuated degrees of freedom.
However, the actions taken by policies $\pi_{\rm source}$ and $\pi_{\rm target}$ may be very different from each other. For example,
the actuators may be calibrated differently,
or realistic actuators may have complex dynamics like fluctuating temperature
or gear backlash, which are not modeled in simulation. The dimensionality of the action space may even be different, for example when the target domain actions may be over biarticular pairs of antagonistic cables or muscle tendons, as in \cite{musclecontrol}. We have such flexibility in our method because the actions generated by
the policy $\pi_{\rm source}$ are never directly used in the target
space, but only through mediation of the simulator and the anticipated
next observation.

\subsection{Training of the inverse dynamics model}
\label{secc:method_training}

We propose to collect trajectories in the physical environment, and to train
a neural network that represents the inverse dynamics model, i.e.,
that can (approximately) predict the action that will lead to the next observation.
For a snippet of a trajectory: $\tau_{H:H+k}$
and next observation $o_{H+k+1}$,
we train a neural network $\phi$ to predict the preceding action $a_{H+k}$:
\begin{align*}
    \phi : (o_H, a_H, o_{H + 1}, \dots, a_{k+H-1}, o_{k+H}, o_{k + H+1}) \mapsto a_{k + H}
\end{align*}
We incorporate history in our model and pick the history window parameter $H$ to be large enough that $\phi$
can (implicitly) infer any important latent factors or temporal dependencies present in the dynamics.

\subsection{Data collection / Exploration}
\label{secc:method_data}

At each point during training
we have a preliminary inverse dynamics model $\phi$,
which we can use to implement a preliminary policy $\pi_{\rm target}$.
In order to collect training data for our model,
we execute this preliminary policy $\pi_{\rm target}$.
We add noise to the prescribed actions for exploration, i.e., in order
to ensure that we have sufficiently diverse training data.
Adding too much noise will result in data collected too far from the target trajectories,
adding too little noise will result in insufficient exploration and
the inverse dynamics model will improve very slowly.  In our
experiments we describe our noise settings.  We found it helpful to \emph{not} add noise at every time step.  Adding noise too 
frequently steers the data collection too far away from the relevant parts of the space for the task at hand.
In simulation we can collect training samples very efficiently by setting the simulator
to the states that occur along a trajectory; in a physical system,
the efficiency of collecting training data depends on the amount of noise that can
be injected into the controls before the robot moves far enough from the target trajectories
that its behavior is no longer useful for training. We also found it more efficient to reset once the target execution starts
deviating very far from what would have happened in the source environment.

\subsection{Inverse dynamics neural network architecture}
\label{secc:method_architecture}

All of our inverse dynamics models $\phi$ take as input a sequence of $k$ previous observations, $k-1$ previous actions,
and a target observation.
Observations and actions are concatenated into one large input vector for the neural net. As is common in current neural net learning practice, the neural network inputs are normalized to have mean $0$ and variance $1$ \cite{alexnet}.
We then apply a sequence of two fully-connected hidden layers with ReLU activations
and $256$ units each,
followed by a fully-connected output layer,
which gives the action $a = \phi(\tau_{-k:}, o)$.

\section{Experiments}

The purpose of our method is to adapt a policy from a source environment to a target environment, with the key application being adaptation from simulation to real world.
First, we measure adaptation capability between two simulators~\ref{secc:sim} as this allows us to quantify most directly the differences between source environment and target environment. 
Then, we present results for adaptation from a simulation to a physical environment.

\begin{figure}[t]
 \centering
 \includegraphics[width=0.24\linewidth]{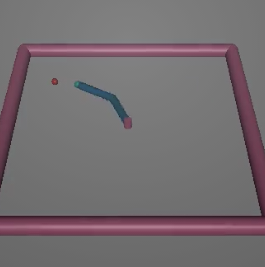}
 \includegraphics[width=0.24\linewidth]{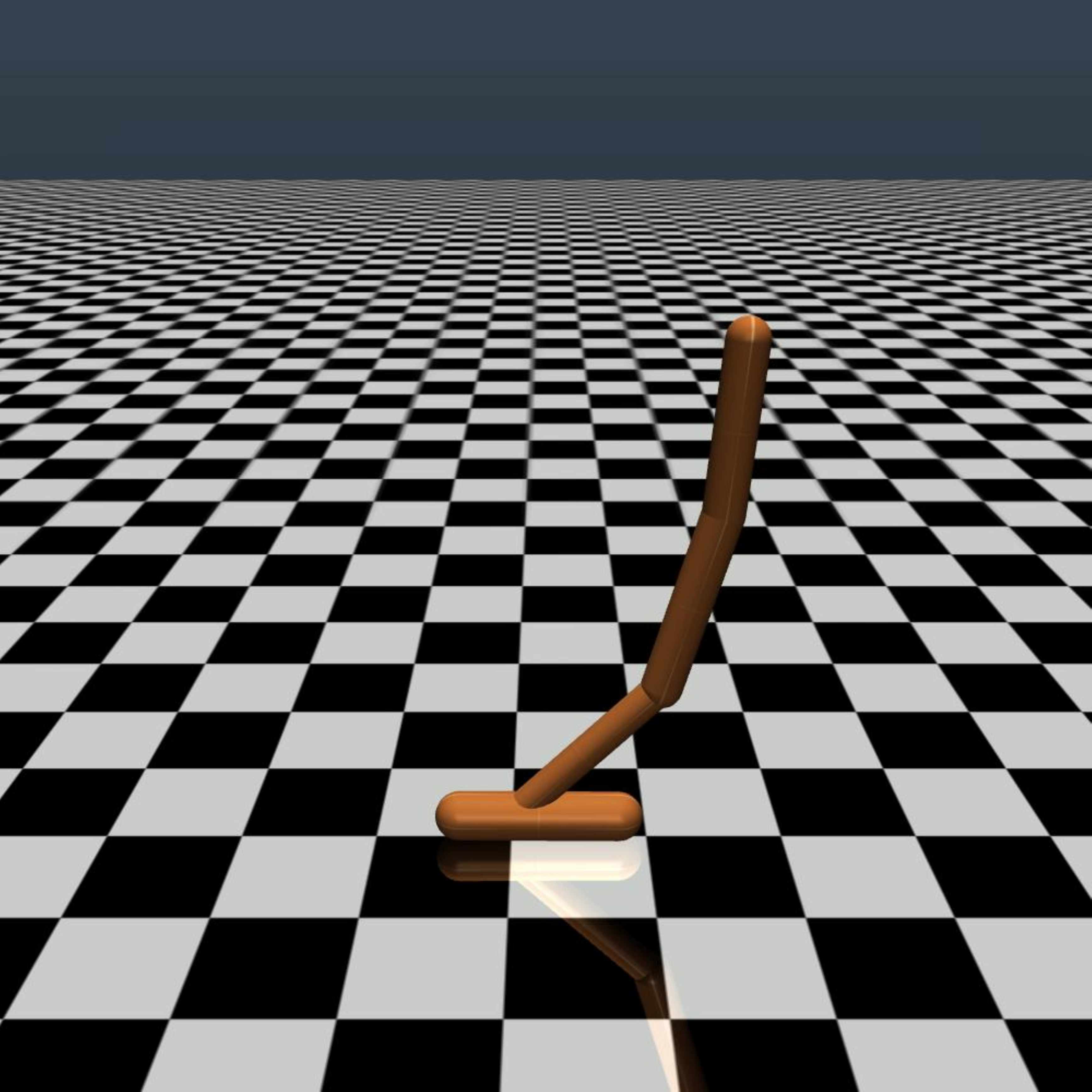}
 \includegraphics[width=0.24\linewidth]{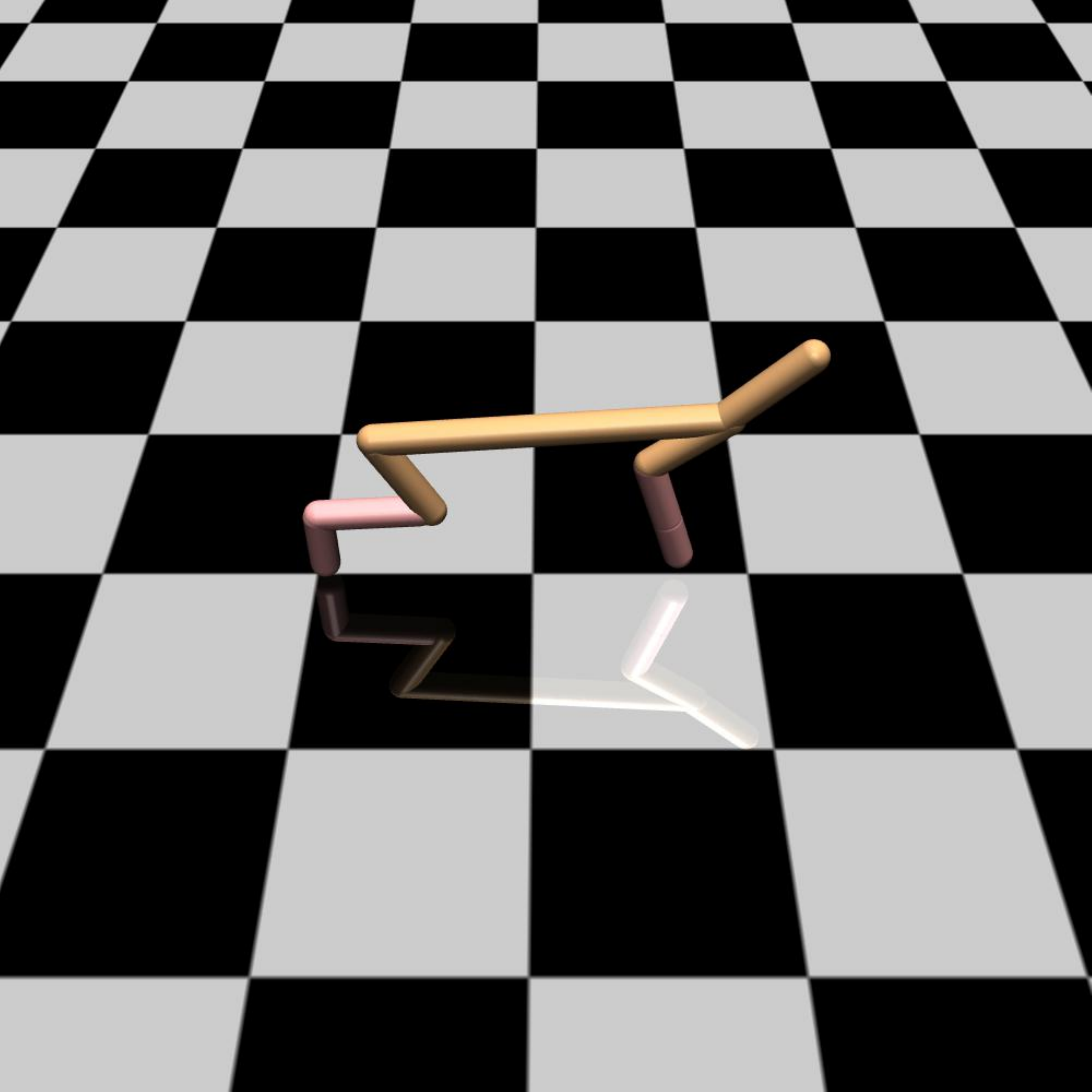}
 \includegraphics[width=0.24\linewidth]{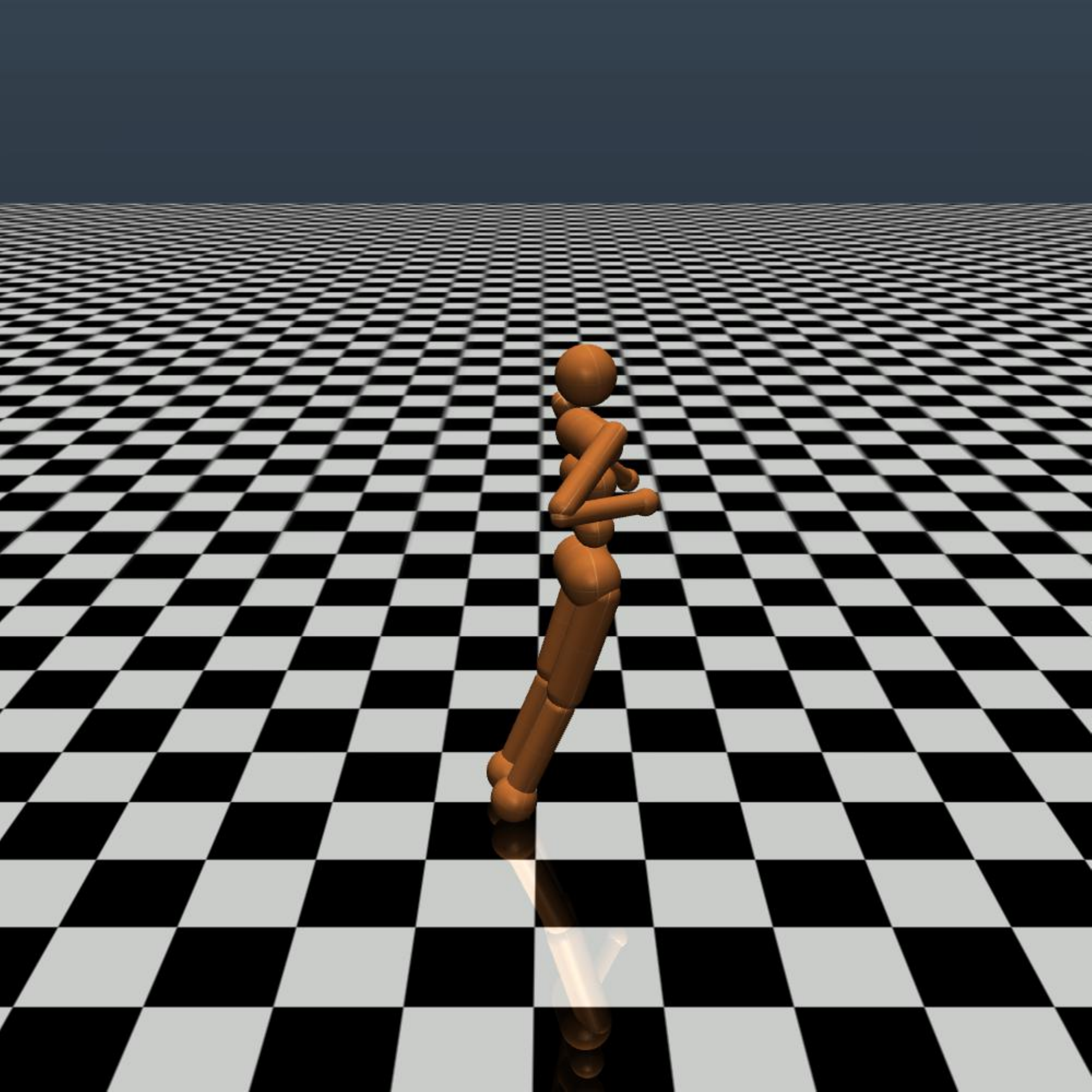}
\caption{Images of environments in our simulation experiments: Reacher, Hopper, Half Cheetah and Humanoid.}
\label{fig:tasks}
\end{figure}

\begin{figure*}[!ht]
  \centering

  \begin{minipage}{0.28\linewidth}
    \centering
    \includegraphics[width=\linewidth]{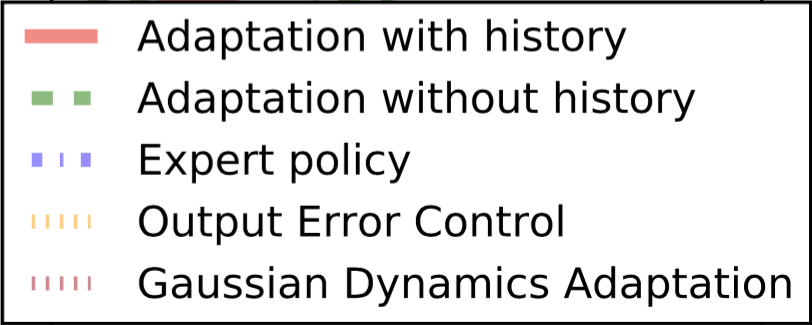} 
  \end{minipage}
  \hfill
  \begin{minipage}{0.68\linewidth}
      \centering
      \rule{\linewidth}{1pt}\\
      {\bf \large Varying gravity}\\
      \vspace{1mm}
        \includegraphics[width=\linewidth]{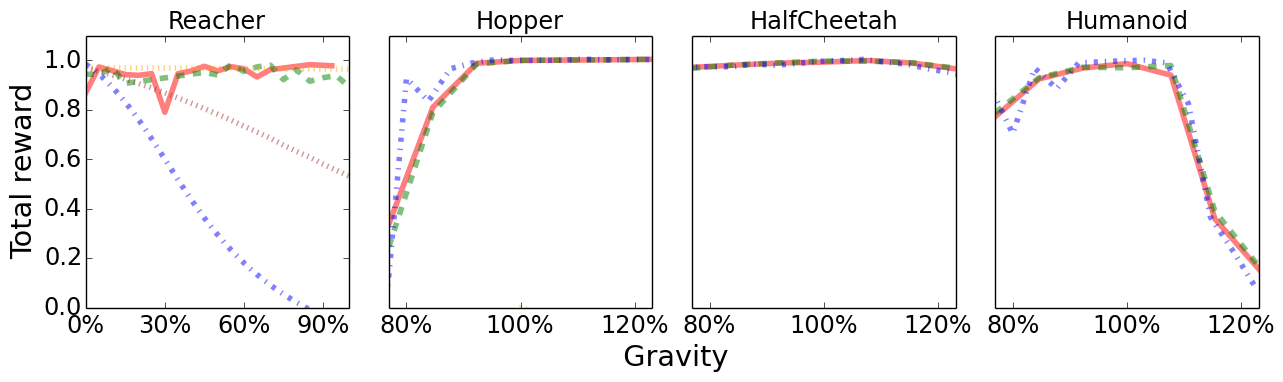} 
  \end{minipage}\\

  \rule{\linewidth}{1pt}\\
  {\bf \large Motor noise}\\
  \vspace{1mm}
  \begin{minipage}{0.47\linewidth}
    \centering
    Reacher\\
    \includegraphics[width=0.99\linewidth]{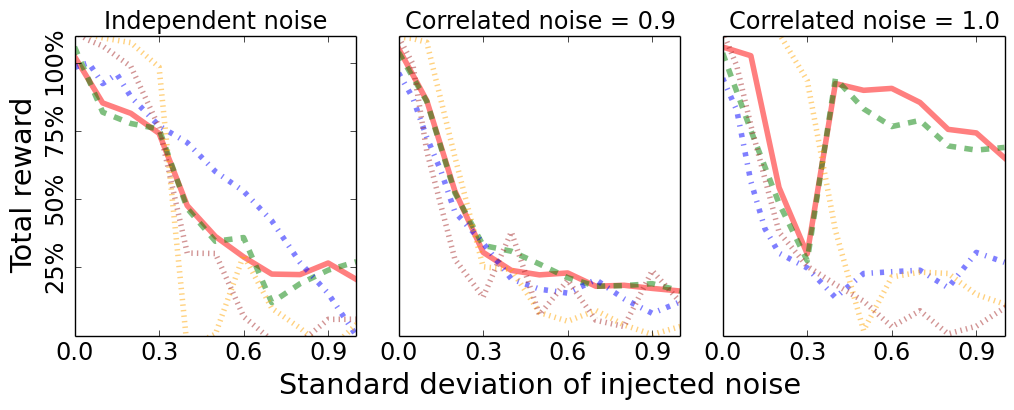} \\
    \hspace{50mm}
  \end{minipage}
  \hfill
  \begin{minipage}{0.47\linewidth}
    \centering
    Hopper \\
    \includegraphics[width=0.99\linewidth]{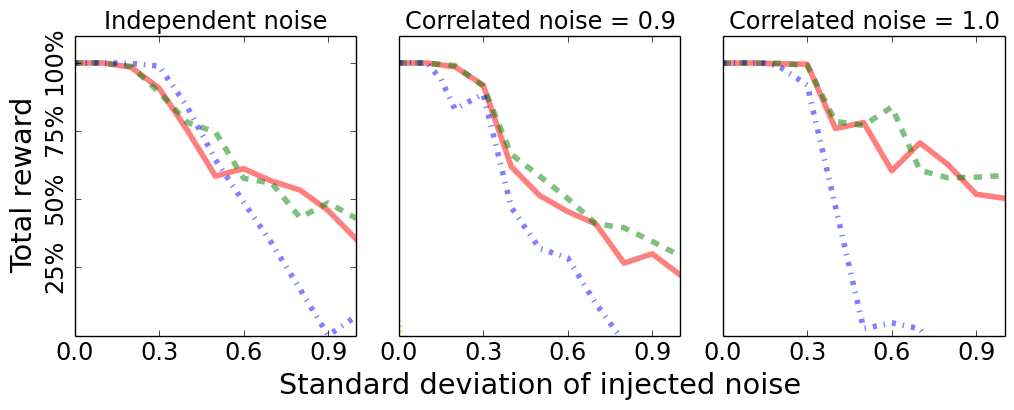} \\
    \hspace{50mm}
  \end{minipage}
  \\
  \begin{minipage}{0.47\linewidth}
    \centering
    Half-Cheetah\\
    \includegraphics[width=0.99\linewidth]{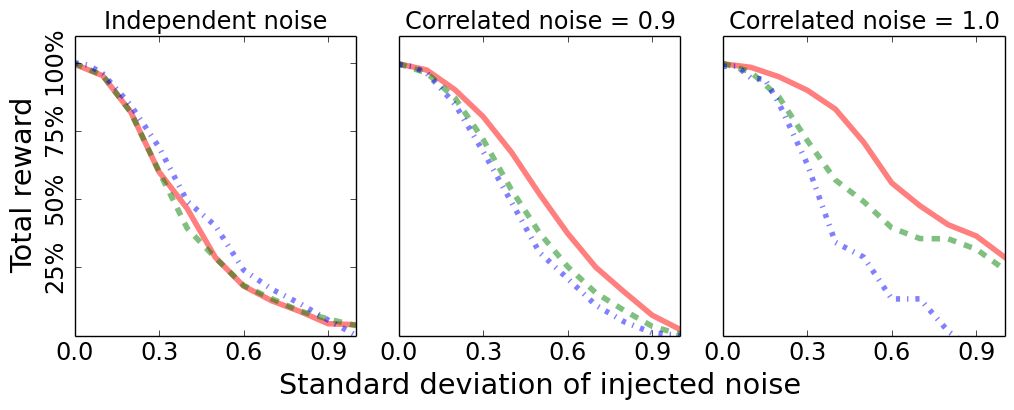} 
    \hspace{50mm}
  \end{minipage}
  \hfill
  \begin{minipage}{0.47\linewidth}
    \centering
    Humanoid\\
    \includegraphics[width=0.99\linewidth]{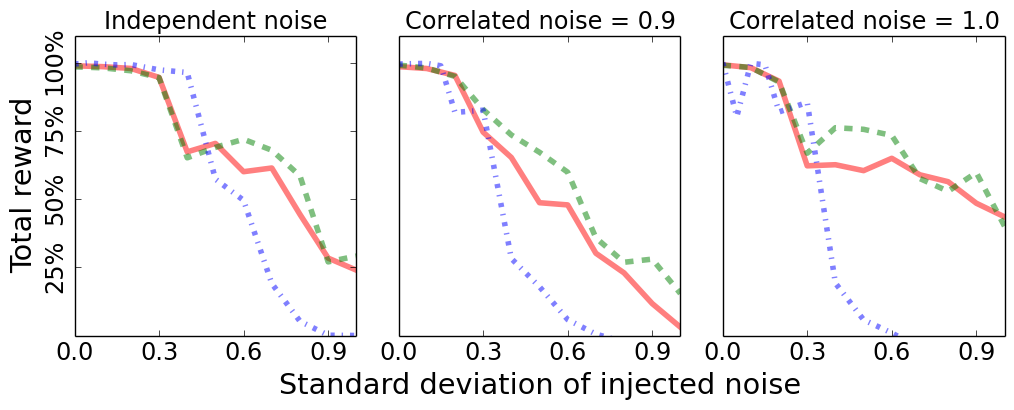} 
    \hspace{50mm}
  \end{minipage}
  \\
  \rule{\linewidth}{1pt}\\
  \caption{Plots present robustness of expert policies and our adaptation method to
           differences between source environment and target environment. x-axis measures
           how much target environment differs from the source environment. 
           y-axis is the normalized cumulative amount of reward averaged over ten random seeds.
           We observe that the
           expert policy performs well in small changes to environment, but is not robust to large changes. Baseline adaptation methods achieve near-zero reward on contact-rich environments. By constrast, our adaptation method performs well in both small and large environment changes - in part due to outputting action correction terms when enviornments are similar. We further gain minor improvement when the adaptation network that is trained with a history windows. }
  \label{fig:resilience}
\end{figure*}

\subsection{Simulated Environments -- Sim1 to Sim2 transfer}
\label{secc:sim}

We test our method on several simulated models in the robotics simulator MuJoCo ~\cite{mujoco} using OpenAI Gym environment ~\cite{brockman2016openai}.
Therefore, both source and target environments are in simulation. We perform experiments on the following standard OpenAI Gym environments (Figure~\ref{fig:tasks}). In each case, observation space consists of positions and velocities of all degrees of freedom.
\begin{itemize}
\item \emph{Reacher.} Two-link arm aiming toward a target location,
with a 11-dimensional observation space and 2 actuators. Arm end effector and target are included in the observation.\footnote{We modify
the Gym environment by increasing the mass of the arm to be $34$ kilograms, roughly in line
with the physical Fetch robot. This has a minimal effect on the original task,
but it becomes relevant when we try to adapt to a modified version of the task with different gravity.}
\item \emph{Hopper.} Two-dimensional model of a robot with a single ``foot''
that moves by hopping, with a 12-dimensional observation space and 3 actuators.
\item \emph{Half-Cheetah.} Two-dimensional model of a bipedal robot
with a 17-dimensional observation space and 6 actuators.
\item \emph{Humanoid.} Three-dimensional model of a humanoid robot
with a 376-dimensional observation space and 17 actuators.
\end{itemize}
In each environment, we train our models to imitate an ``expert policy''.
The expert policies are obtained from Trust Region Policy Optimization \cite{schulman2015trust} (source code by Ho et. al~\cite{ho2016generative}).
We measure the performance of policies using a reward given by OpenAI gym~\cite{brockman2016openai}.\footnote{These reward functions
feature penalties for applying large torques; we remove these penalties,
because they make it more difficult to interpret results which require gravity compensation
or for which there is motor noise.}
We normalize the performance measurement so that the performance of the expert policy
is $1$.

Note that our algorithm never observes the performance of the adapted policy.
This is important for our intended application;
evaluating the performance of adapted policies operating in the real world is typically
more expensive than executing those policies, as it might, for example, require instrumentation of the physical world with ground truth sensors.
We only use the performance measures to determine whether our method
has successfully adapted the critical features of the expert policy.

To produce training data,
we interleave learning with execution in the target domain, executing the previous estimate of the inverse dynamics model $\phi$ to
generate trajectories to be used for further training.
We interrupt trajectories at a random point
in order to take a random action,
and train the model to predict the random action from the resulting state
(as well as the history of recent states and actions).
We report all of our training times in terms of the number of training samples that we collect. In the case when inverse dynamics model includes history (as described in ~\ref{secc:method_training}), we use a window size $H=2$ for all the experiments.

We compare our approach to several popular methods that have been developed to deal with simulation to real world model discrepancy. The baselines we use are:
\begin{itemize}
\item \emph{Expert Policy.} We perform no adaptation and directly use the actions of the policy obtained from source domain in the new target domain. $a{\rm target} = \pi_{\rm source}$.  

\item \emph{Output Error Control.} We perform Model Predictive Control in the target domain using an adapted version of a dynamics model $T_{\text{source}}$ transferred from the source domain. 
At each timestep, we use the current observation and previous action to update the dynamics model, and use the updated dynamics model to compute a policy using iterative LQR \cite{todorov05}. Output Error Control dynamics adaptation scheme adjusts the source dynamics model
$$T_{\text{target}}^t = T_{\text{source}} + e_t$$ 
by an error term 
$$e_t = (1 - \gamma) e_{t-1} + \gamma (o_t - T_{\text{source}}(o_{t-1}, a_{t-1}))$$
representing a decayed version of the error in $T_{\text{source}}$ in the target domain.

\item \emph{Gaussian Dynamics Adaptation.} As the previous baseline, we perform Model Predictive Control using iterative LQR on an adjusted dynamics model. The adjustment scheme in this case uses the source dynamics model to form a local Gaussian prior 
$p(o_{t}, a_{t}, o_{t+1}).$
We update this prior according to the empirical mean and covariance of the data observed in target domain, and condition it to form 
$$T_{\text{target}}^t = p(o_{t+1} \mid o_t, a_t).$$
This is the approach proposed and described in more detail in \cite{fu15}.
\end{itemize}

To test the capability of our method compared to the baseline methods,
we consider two following challenging differences between domains:
\begin{itemize}
\item \emph{Variation in Gravity.} Target environment has a difference in gravity from the source environment. Gravity differs in magnitude by $20\%$ for locomotion tasks.
The Reacher task occurs in a plane; the expert policy is trained in a horizontal plane and essentially unaffected by gravity,
and we test on planes that are rotated from $0^\circ$ to $90^\circ$.
On the Reacher task, our method is able to adapt successfully to this significant dynamics change.
\item \emph{Motor Noise.} Before an action $a$ is sent to the robot,
it is perturbed by adding a noise term to obtain $a' = a + \epsilon_t$.
We experiment with two variants, where this noise is independent on each time step,
as well as where this noise varies slowly and is correlated over time. Such noise is more representative of fluctuating environmental conditions, or latent physical effects like temperature changes.
\end{itemize}

In many cases, only small corrections to the source domain actions are necessary to adapt to target domain. In such a setting, it may be beneficial for $\phi$ to output a correction term rather than an action directly:
$$a_{\rm target} = a_{\rm source} + \phi(\tau_{-k:}, \hat{o}_{\rm next}).$$
This has the downside of directly requiring actions from the source domain, but tends to result in better performance when the domains are similar. We use such a correction formulation for motor noise standard deviations below 0.3 and for all locomotion experiments with varying gravity. In such cases, we also found it most helpful to pre-train the model on trajectories produced by the expert policy.

Figure~\ref{fig:resilience} summarizes our results, and
Table \ref{tab:sample_complexity} presents sample complexity  of our method.

As expected, simply applying actions from an expert policy from a source domain results in poor performance on the target domain. Baselines that perform planning using a locally Gaussian forward dynamics model that is adapted online performed well with no additional training on the target domain in environments with simple dynamics (e.g., no contacts) such as Reacher and relatively slowly changing variation between the source and target domain. However, we found these methods to be ineffective in contact-rich environments such as Hopper, Cheetah, and Humanoid, even in the source domain. Contacts induce discontinuities that cause methods using locally linear dynamics approximations to perform poorly. Unstable tasks like Hopper and Humanoid are particularly poorly suited for these methods because small errors propogate over long trajectories, leading to episode termination.

Our method is also able to correct for slowly-varying noise and small changes to system dynamics. Moreover, it is able to adapt even in the presence of contact discontinuities that are extermeley challenging for approaches based Model Predictive Control. Such approaches require solving an optimization problem (iterative LQG) that can exploit the learned forward dynamics model and take it outside the regime it was trained on. By learning an inverse dynamics model, we simply take the output of such models and avoid performing potentially unstable numerical optimization.

\begin{figure}[h!]
\scriptsize
\centering
\renewcommand{\arraystretch}{1.3}
\begin{tabular}{|c|c|c|c|c|c|c|c|}
\hline 
Noise std. & none & \multicolumn{3}{|c|}{0.2}      & \multicolumn{3}{|c|}{1.0}    \\ 
\cline{3-8}
Noise correlation        & none & 0.0 & 0.9 & 1.0  & 0.0 & 0.9 & 1.0 \\ \hline
\multicolumn{8}{|c|}{{\bf Number of training samples in thousands (smaller is better)}} \\ \hline

 \multicolumn{8}{|c|}{{\bf Hopper }} \\ \hline
Adaptation without history & 31 & 58 & 48 & 77 & 150 & 157 & 137\\ \hline
Adaptation with history & 24 & 31 & 29 & 28 & 70 & 121 & 47\\ \hline
Learning from scratch & \multicolumn{7}{|c|}{{ about 1000 }} \\ \hline

 \multicolumn{8}{|c|}{{\bf Humanoid  }} \\ \hline
Adaptation without history & 13 & 15 & 20 & 16 & N/A & N/A & 155\\ \hline
Adaptation with history & 16 & 17 & 19 & 16 & N/A & N/A & 54\\ \hline
Learning from scratch & \multicolumn{7}{|c|}{{ about 70000 }} \\ \hline

\end{tabular}
\caption{Table presents the number of samples required to converge to 3/4 of the expert policy's performance,
running on our simulated environments with additive noise. The method requires more samples in the presence
of noise. Including history typically reduces the complexity of the learning problem.
As a comparison, the fastest RL algorithms posted to OpenAI Gym as of submission
require 70 million samples to converge to 75\% performance on Humanoid, 
and about a million samples to converge to 75\% performance on Hopper,
so these running times are about two orders of magnitude faster than those required
to learn policies directly.
}
\label{tab:sample_complexity}
\end{figure}

\subsection{Physical interaction -- Sim to Real transfer}
We test our method on transferring trajectories from a simulated source domain 
to the target domain, which is physical Fetch robot \cite{fetch}.
We control the robot using position control and stock firmware based on ROS in 10Hz frequency.

The tasks consider control of the arm, and our metric measures normalized distance between 
observations achieved in the simulator by the trajectory and observations achieved on the physical robot. 
The task is an agile back-and-forth swing of an arm where middle of the arm is pulled by a bungee cord. 
Our action adaptation method is able to adjust to this condition by adapting and exerting the necessary about of torque.
As a baseline we use PD controller with targets being 
states experienced in the simulator.
Table~\ref{tab:sample_complexity} summarizes
our results.

\begin{figure}[h!]
\scriptsize
\centering
\renewcommand{\arraystretch}{1.3}
\begin{tabular}{|c||c|}
\hline 
    \diagbox{Method}{Task} & Swings limited with a bungee cord \\ \hline \hline
    Our method    & $3.72 \% \pm 0.020 \% $  \\ \hline
    PD controller & $4.49 \% \pm 0.050 \% $  \\ \hline
\end{tabular}
\caption{Table presents average distance and varaince to the desired trajectory for our method and
PD baseline, averaged over 10 trials. }
\label{tab:sample_complexity}
\end{figure}

\section{Discussion and Future Work}

We have presented a general method to adapt actions of policies developed in one domain such as simulation to a different domain such as the physical world. We achieve this by learning a deep inverse dynamics model that is trained on the behavior of the physical robot. Our method is successfully able to adapt complex control policies for aggressive reaching and locomotion on scenarios involving contact, hysteresis effects in the form of time-correlated noise, and significant differences between environments. However to bring about robots that truly generalize in the physical world, in addition to action adaptation it is necessary to also adapt states and observations between simulation and physical world. We currently assume observations generated by our simulator match closely to physical observations, which is reasonable when considering sensors such as joint positions, but is it not reasonable to expect simulated visual or depth sensors to match the high fidelity of the real world. This work only focused on action adaptation. In the future we plan to experiment with observation adaptation methods, such as \cite{tzeng15} for instance. Additionally, our approach can be applied to a setting where we do not even observe the actions taken in the source domain. This presents exciting future opportunities to apply our method to use solely observations in the source domain (such as driving dashboard camera recording, for example) to recover and adapt actions for a corresponding driving policy.

\bibliographystyle{plain}
\bibliography{refs}

\end{document}